\documentclass[10pt,twocolumn,letterpaper]{article}

\usepackage{cvpr}              % To produce the CAMERA-READY version
\definecolor{cvprblue}{rgb}{0.21,0.49,0.74}
\usepackage[pagebackref,breaklinks,colorlinks,allcolors=cvprblue]{hyperref}

\usepackage{xcolor}
\usepackage{multirow}
\usepackage{colortbl}
\usepackage{adjustbox}
\usepackage{pifont}
\usepackage{makecell}
\usepackage{array}
\usepackage{tikz}
\usepackage{amssymb}
\usepackage{mathabx}
\usepackage{fontawesome}

\definecolor{my_green}{RGB}{0, 255, 0}
\definecolor{my_red}{RGB}{255, 0, 0}

\definecolor{figgreen}{RGB}{59, 125, 35}
\definecolor{figorange}{RGB}{215, 137, 0}
\definecolor{figred}{RGB}{192, 0, 0}
\definecolor{figblue}{RGB}{48, 132, 194}
\definecolor{mygray}{gray}{0.8}
\definecolor{Red}{rgb}{0.8,0,0}
\definecolor{Blue}{rgb}{0,0,0.8}
\definecolor{gg}{gray}{0.92}
\definecolor{tabred}{RGB}{230, 10, 10}
\definecolor{tabblue}{RGB}{10, 10, 230}
\definecolor{citeblue}{rgb}{0.15,0.35,0.55}
\definecolor{linkred}{rgb}{0.9,0,0}
\definecolor{Blue9}{rgb}{0.1,0.3,0.95}

\usepackage{tcolorbox}
\tcbuselibrary{skins} 
\tcbuselibrary{listings}
\tcbuselibrary{breakable}

\newtcblisting{promptbox}[1][]{
    enhanced,
    attach boxed title to top left={xshift=1em, yshift=-\tcboxedtitleheight/2},
    colback=Blue9!5!white,
    colframe=Blue9!50!black,
    colbacktitle=Blue9!50!black,
    coltitle=white,
    listing only,
    boxed title style={sharp corners, frame hidden},
    fonttitle=\ttfamily\centering\bfseries,
    title={#1},            
    listing options={
        language={},
        basicstyle=\tiny\ttfamily,
        breaklines=true,
        breakautoindent=false,
        breakindent=0pt,
        columns=fullflexible,
        showstringspaces=false,
    },
    boxsep=2pt,
    top=2pt,
    bottom=2pt,
}

\makeatletter
\newcommand{\thickhline}{%
 \noalign {\ifnum 0=`}\fi \hrule height 1pt
 \futurelet \reserved@a \@xhline
}
\makeatother

\def\paperID{18824} % *** Enter the Paper ID here
\def\confName{CVPR}
\def\confYear{2026}

\title{MA-Bench: Towards Fine-grained Micro-Action Understanding}

\author{Kun Li$^1$, Jihao Gu$^2$, Fei Wang$^{3,4}$, Zhiliang Wu$^5$, Hehe Fan$^5$, Dan Guo$^{3,4}$\thanks{Corresponding author.} \\
\small  \normalsize$^1$ CVLab, College of Information Technology, United Arab Emirates University  
\small \normalsize$^2$ University College London\\ 
\small \normalsize$^3$ Hefei University of Technology 
\small \normalsize$^{4}$ Institute of Artificial Intelligence, Hefei Comprehensive National Science Center \\
\small \normalsize$^{5}$ 
CCAI, Zhejiang University 
}

\begin{document}
 
\maketitle

\begin{abstract}
With the rapid development of Multimodal Large Language Models (MLLMs), their potential in Micro-Action understanding, a vital role in human emotion analysis, remains unexplored due to the absence of specialized benchmarks. 
To tackle this issue, we present \textbf{MA-Bench}, a benchmark comprising 1,000 videos and a three-tier evaluation architecture that progressively examines micro-action perception, relational comprehension, and interpretive reasoning.
MA-Bench contains 12,000 structured question–answer pairs, enabling systematic assessment of both recognition accuracy and action interpretation. 
The results of 23 representative MLLMs reveal that there are significant challenges in capturing motion granularity and fine-grained body-part dynamics. 
To address these challenges, we further construct \textbf{MA-Bench-Train}, a large-scale training corpus with 20.5K videos annotated with structured micro-action captions for fine-tuning MLLMs. 
The results of Qwen3-VL-8B fine-tuned on MA-Bench-Train show clear performance improvements across micro-action reasoning and explanation tasks. 
Our work aims to establish a foundation benchmark for advancing MLLMs in understanding subtle micro-action and human-related behaviors. Project Page: \href{https://MA-Bench.github.io}{https://MA-Bench.github.io}.
\end{abstract}

\section{Introduction}
\label{sec:intro}

Human body actions, as an important form of non-verbal communication, effectively convey emotional information in social interactions~\cite{aviezer2012body}. In this paper, we investigate the subset of human body actions, namely Micro-Actions (MAs)~\cite{liu2021imigue,chen2023smg,guo2024benchmarking,liu2024micro,li2025prototypical,li2025mmad,gu2025motion}, which reflect spontaneous body movements caused by emotional changes. Micro-Action analysis~\cite{chen2023smg,li2025prototypical,guo2024benchmarking,li2025mmad,guo2024mac,li2025mac,li2023data,shang2025cross} has emerged as a significant research area due to its potential applications in human-to-human communication and human emotional state analysis~\cite{zhao2025temporal,wang2026xinsight}.

\begin{figure*}
\centering
\includegraphics[width=\textwidth]{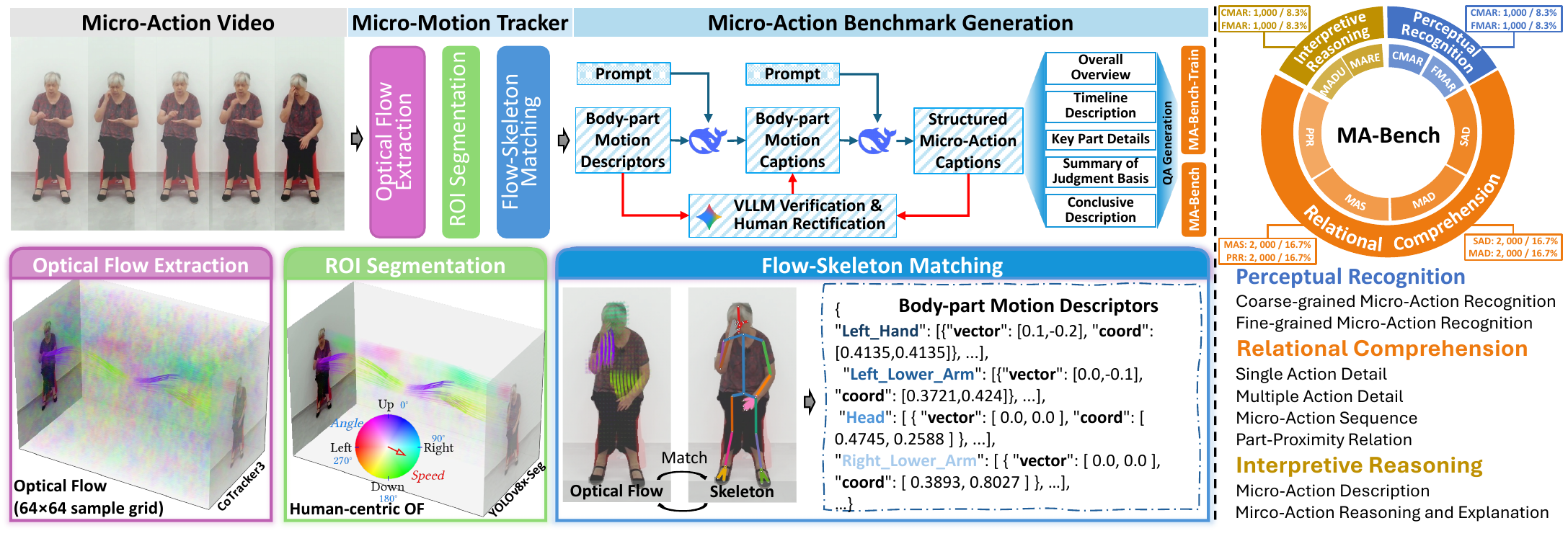}
\vspace{-1.4em}
\caption{
The pipeline for constructing \textbf{MA-Bench} and \textbf{MA-Bench-Train} for \textbf{fine-grained micro-action understanding}. (1) \textbf{Micro-motion tracker} extracts motion descriptors (\ie, motion vectors and coordinates) for each body part. (2) \textbf{Micro-action benchmark generation} leverages motion descriptors and multimodal large language models to create structured micro-action captions, which are then used to generate the benchmarks. (3) \textbf{MA-Bench} enables fine-grained action understanding through well-defined \textbf{perceptual recognition}, \textbf{relational comprehension}, and \textbf{interpretive reasoning}.
}
\label{fig:teaser}
\end{figure*}

Micro-action analysis~\cite{liu2021imigue,chen2023smg,guo2024benchmarking,li2025mmad} is still in its infancy. To advance the study of these subtle movements, several datasets have been developed. 
Liu~\etal~\cite{liu2021imigue} collected the identity-free video dataset for Micro-Gesture Understanding and Emotion analysis (iMiGUE). iMiGUE focuses on micro-gestures that are unintentional behaviors driven by inner feelings. It consists of 18, 499 samples across 32 categories, collected from the post-matching interview scenario. Similarly, Chen~\etal~\cite{chen2023smg} collected the Spontaneous Micro-Gestures (SMG) dataset that was collected under objective proxy tasks to stimulate two states of emotional stress (\ie, positive and negative). The SMG dataset consists of 414 long video instances and 3,712 micro-gesture clips spanning 16 categories. 
In contrast, Balazia~\etal~\cite{balazia2022bodily} proposed the BBSI dataset that captured the complex bodily behaviors (\eg, ``adjusting clothing'' and ``fold arms'') exhibited by three to four individuals engaged in spontaneous group discussions. BBSI is a multi-view dataset comprising 26 hours of video annotated with 15 bodily behavior classes. 
However, these datasets are mainly focused on the actions occurring in the upper limbs. To explore the full-body actions for micro-action analysis, Guo~\etal~\cite{guo2024benchmarking} collected the large-scale micro-action recognition dataset Micro-Action-52, which consists of 22, 422 video instances collated from the psychological interviews. These micro-actions are annotated with 7 body-level categories (\ie, ``body'', ``head'', ``upper limb'', ``lower limb'', ``body-hand'', ``head-hand'' and ``leg-hand''), and 52 action-level categories (\eg, ``shaking body'', and ``turning head''). 

\begin{table}[t!]
\caption{Comparison of MA-Bench with existing video understanding benchmarks. }
\tabcolsep 2pt
\resizebox{1.0\linewidth}{!}{
\begin{tabular}{|r|c|c|c|c|}
\hline\thickhline
\rowcolor{mygray} Benchmark & \#Videos & Topic  & \#QA & \#Annotation \\ \hline
MSVD-QA~\cite{xu2017video} & 520 & Daily Activities    & 13,157 & Auto \\
MSRVTT-QA~\cite{xu2017video} & 800  & Daily Activities   & 8,000 & Auto \\
ActivityNet-QA~\cite{yu2019activitynet}  & 2,990  & Daily Activities   & 72,821 & Manual \\
NExT-QA~\cite{xiao2021next} & 1,000  & Daily Activities    & 8, 564 & Manual \\
Video-MME~\cite{fu2025video} & 900 & Daily Activities    & 2,700  & Manual \\
EgoTaskQA~\cite{jia2022egotaskqa} & 2,315 & Ego Activities & 40,322  & Manual \\
EgoSchema~\cite{mangalam2023egoschema} & 5,031 & Ego Activities   & 5,031 & Auto\&Manual \\
MVBench~\cite{li2024mvbench}   & 4,000 & Long Videos   & 4,000 & Auto\&Manual \\
TVBench~\cite{cores2024tvbench}  & 2,525 & Long Videos   & 2,525 & Manual \\
MLVU~\cite{zhou2024mlvu}  & 1,730 & Long Videos   & 3,102 & Auto\&Manual \\
LongVideoBench~\cite{wu2024longvideobench}  & 3,763 & Long Videos & 6, 678 & Manual\\
LVBench~\cite{wang2025lvbench} &  103  & Long Videos & 1,549 &  Manual \\
MovieQA~\cite{makarand2015movieqa} & 408 & Movie &  6,462 & Manual \\
MovieChat-1K~\cite{song2024moviechat} & 130   & Movie   & 1,950 & Manual \\
MotionBench~\cite{hong2025motionbench}  & 5,385 & Fine-grained Motion & 8,052 & Auto\&Manual \\
FAVOR-Bench~\cite{tu2025favor}  & 1,776 & Fine-grained Motion & 8,184 & Manual \\ \hline
\rowcolor{mygray} \textbf{MA-Bench (Ours)} & 1,000 & Fine-grained Micro-Action    & 12, 000 & Auto\&Manual \\ \hline   
\end{tabular}}
\label{tab:dataset_cmp}
\end{table}

Driven by these datasets, numerous approaches~\cite{liu2021imigue,guo2024benchmarking,li2025prototypical,gu2025motion} have been developed to address challenges such as data imbalance, action-label alignment, and ambiguity in recognition. For example, Liu~\etal~\cite{liu2021imigue} tackled the imbalanced data issue in in-the-wild settings with a skeleton-based unsupervised network. Guo~\etal~\cite{guo2024benchmarking} proposed a network guided by label semantic constrained loss to improve the alignment of actions with their labels. Similarly, Li~\etal~\cite{li2024enhancing} introduced a visual-text contrastive learning framework that leverages textual information from action labels for enhanced micro-gesture recognition. In contrast, Li~\etal~\cite{li2025prototypical} proposed a prototypical calibrating network to reduce ambiguity in micro-action recognition through hierarchical prototypes and contrastive calibration. Recently, Gu~\etal~\cite{gu2025motion} present a motion-guided modulation network that utilizes skeletal- and frame-level motion cues to improve spatial-temporal representation learning.

In recent years, MLLMs~\cite{team2023gemini, wu2024next,zhu2025internvl3,hurst2024gpt,yang2024qwen2.5,qwen3technicalreport} have rapidly gained prominence as a new direction of research. 
Several datasets~\cite{fu2025video,li2024mvbench,hong2025motionbench,tu2025favor} have been introduced to facilitate MLLM-based studies in video understanding through visual question answering, enabling models to align visual dynamics with textual semantics. 
As shown in Table~\ref{tab:dataset_cmp}, MSVD-QA~\cite{xu2017video}, MSRVTT-QA~\cite{xu2017video}, and ActivityNet-QA~\cite{yu2019activitynet} are focused on descriptive QA in daily activities. In contrast, NExT-QA~\cite{xiao2021next} focuses on the causal temporal action interactions in videos. Video-MME~\cite{fu2025video} comprehensively evaluates MLLMs across various video-related tasks. 
Recently, there have been various benchmarks proposed for different scenarios, such as ego activities~\cite{jia2022egotaskqa,mangalam2023egoschema}, long videos~\cite{li2024mvbench,wu2024longvideobench,wang2025lvbench}, movie~\cite{makarand2015movieqa,song2024moviechat}, and fine-grained motion~\cite{hong2025motionbench,tu2025favor}. 
However, despite the critical importance of micro-actions in emotion analysis and affective computing~\cite{liu2021imigue,chen2023smg,guo2024benchmarking}, there remains a lack of dedicated datasets in this domain, leaving fine-grained micro-action understanding underexplored in the era of MLLMs. 

To address this gap, we introduce the \textbf{MA-Bench} (\textbf{M}icro-\textbf{A}ction \textbf{Benchmark}) and MA-Bench-Train for evaluating the capacity of MLLMs in fine-grained micro-action understanding. 
As shown in Figure~\ref{fig:teaser}, the pipeline for constructing MA-Bench and MA-Bench-Train consists of three main stages. First, the micro-motion tracker extracts motion descriptors for each body part. 
Next, these motion descriptors and prompts are fed into multimodal large language models to create structured micro-action captions. 
Finally, these captions are used to generate the MA-Bench and MA-Bench-Train. 
The proposed MA-Bench realizes the fine-grained micro-action understanding through three well-structured categories, \ie, perceptual recognition, relational comprehension, and interpretive reasoning. These details are given in Section~\ref{sec:bench}. 
In summary, the contributions of this paper are as follows:
\begin{itemize}
\item We propose a novel annotation strategy for fine-grained micro-action understanding, which integrates optical flow and skeleton information to construct motion descriptors for each body part.

\item We collect a comprehensive benchmark, named the \textbf{MA-Bench} (\textbf{M}icro-\textbf{A}ction \textbf{Benchmark}), designed to evaluate multimodal large language models in the domain of fine-grained micro-action understanding.

\item We introduce MA-Bench-Train, a large-scale training corpus consisting of 20.5K videos with structured micro-action captions, providing high-quality supervision for model training and fine-grained motion understanding.

\item We conduct extensive evaluations on 23 multimodal large language models, including both proprietary and open-source ones, using the proposed MA-Bench. The results reveal that current MLLMs struggle to capture subtle motion patterns and fine-grained temporal dynamics.
\end{itemize}

%%%%%%%%%%%%%%%%%%%%%%%%%%%%%%%%%%%
\section{Related Work}
\subsection{Micro-Action Understanding}
Micro-Action understanding~\cite{guo2024benchmarking,liu2021imigue,chen2023smg,li2025mmad,li2023joint,gu2025mm,liu2025online} focuses on identifying subtle, low-intensity body movements that reflect spontaneous emotional changes. We review related work from the perspectives of datasets and methods.
From a dataset perspective, the iMiGUE dataset~\cite{liu2021imigue} was collected from 72 athletes during post-sports press conferences and covered 32 micro-gesture categories with identity-free, high-quality videos. The SMG dataset~\cite{chen2023smg} was later introduced to support both recognition and detection tasks, consisting of video from 40 participants narrating fabricated or real stories, during which subtle gestures and corresponding emotional states were annotated. 
The Bodily Behaviors in Social Interaction dataset~\cite{balazia2022bodily} adopted a naturalistic multi-view group conversation setting to investigate body language in social contexts such as leadership and rapport. It captures 15 distinct body movements (\eg, ``Adjusting clothing'', ``Hand-mouth'', and ``Leg movement''). However, these datasets are predominantly toward upper-limb micro-actions. To capture more fine-grained variations, Guo~\etal~\cite{guo2024benchmarking} collected the first human-centered whole-body micro-action dataset named Micr-Action 52 (MA-52). Specifically, MA-52 comprises 22K samples across 52 action-level and 7 body-level categories collected from professional psychological face-to-face interviews. 

From a methodological perspective, Liu~\etal~\cite{liu2021imigue} introduced an unsupervised encoder–decoder network that employs KL divergence to guide the model in learning intrinsic action and gesture representations for micro-gesture recognition. Huang~\etal~\cite{huang2023micro} proposed an ensemble hypergraph-convolution Transformer framework that integrates hypergraph-based attention into Transformers to better capture subtle temporal cues to address the long-tailed distribution of categories. 
Based on this, Huang~\etal~\cite{huang2024multi} proposed a multi-scale heterogeneous ensemble network that integrates multi-modal data, combines two complementary architectures with multi-scale residual connections to capture fine-grained features. 
Guo~\etal~\cite{guo2024benchmarking} proposed a network that utilizes a joint-embedding loss function to constrain the semantic distance between video data and action labels. 
To address the inherent ambiguity of micro-actions, Li~\etal~\cite{li2025prototypical} proposed a prototypical ambiguous calibrating network that first constructed body-level and action-level prototypes and then calibrated ambiguous samples by regulating their distances to the corresponding prototypes. 
More recently, Gu~\etal~\cite{gu2025motion} developed a motion-guided modulation network, motivated by the observation that skeleton sequences, compared with RGB data, provide a compact and noise-resilient representation. 

\subsection{MLLMs for Video Understanding}
Multimodal large language models (MLLMs)~\cite{cheng2024videollama2,zhang2025videollama3,wang2024qwen2,qwen3technicalreport} have become an important direction in video understanding~\cite{li2023videochat,zhang2023video,li2025repetitive,zhu2025exploiting,li2023stprivacy}. 
On the benchmarking side, MVBench~\cite{li2024mvbench} spans 20 capabilities from perception to reasoning and is widely used to diagnose weaknesses in temporal and causal understanding. Video-MME~\cite{fu2025video} provides a broader coverage, evaluating across diverse domains, durations, and modalities. Additionally, there are numerous benchmarks for different tasks, such as ego video understanding~\cite{jia2022egotaskqa,mangalam2023egoschema}, long video understanding~\cite{li2024mvbench,wu2024longvideobench,wang2025lvbench}, and fine-grained motion understanding~\cite{hong2025motionbench,tu2025favor}. 
From the methodological perspective, Video-LLaMA~\cite{zhang2023video} introduced joint audio–visual encoding with instruction tuning, enabling LLMs to process video and audio streams. Video-LLaVA~\cite{lin2023video} unified image and video representations through an ``align-before-project'' strategy, achieving stronger generalization via mixed image–video instruction tuning. 
VideoChat2~\cite{li2024mvbench} further demonstrated the effectiveness of progressive multimodal instruction tuning.
Recently, reinforcement learning has been introduced to video understanding to advance reasoning capabilities, such as Video-R1~\cite{feng2025video} and VideoRFT~\cite{wang2025videorft}. 
\textit{However}, existing benchmarks and methods are not designed for the unique challenges of micro-action recognition. In this work, we introduce MA-Bench and MA-Bench-Train to systematically evaluate and advance MLLMs in fine-grained micro-action understanding.

\begin{figure*}[t!]
\centering
\begin{subfigure}[t]{0.51\linewidth}
\centering
\includegraphics[width=1.0\linewidth]{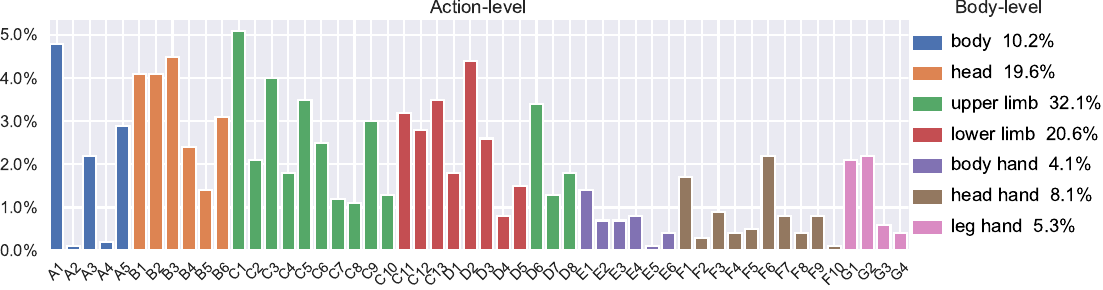}
\caption{The distribution of micro-action categories.}
\end{subfigure}
\begin{subfigure}[t]{0.29\linewidth}
\centering
\includegraphics[width=1.0\linewidth]{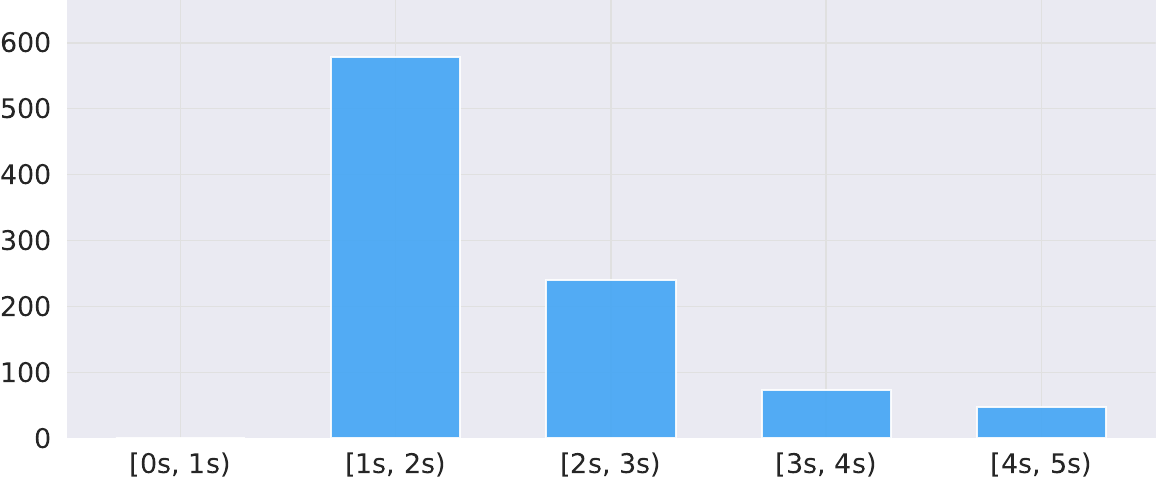}
\caption{The duration of videos}
\end{subfigure}
\begin{subfigure}[t]{0.19\linewidth}
\centering
\includegraphics[width=1.0\linewidth]{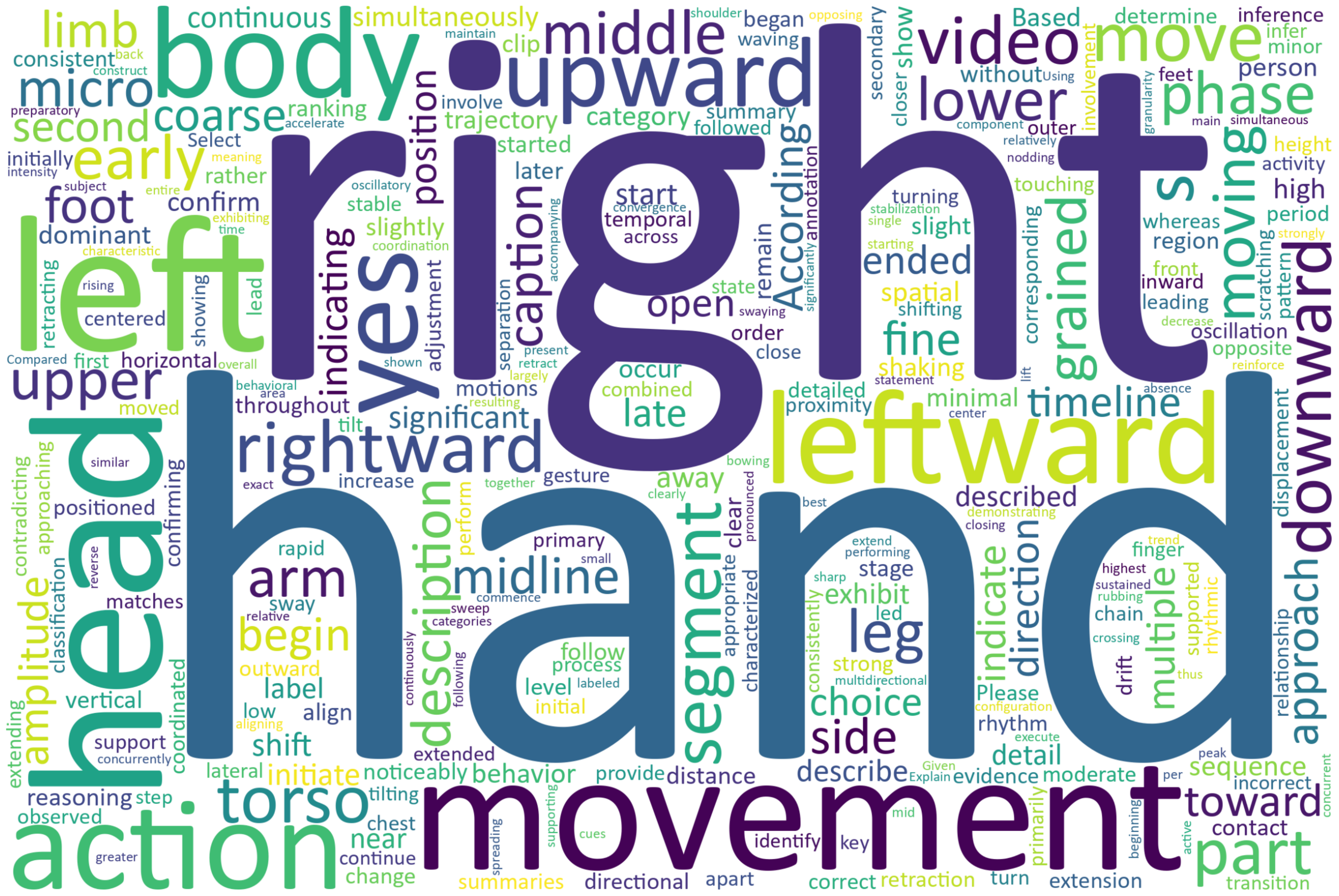}
\caption{Word Cloud Statistics.}
\end{subfigure}
\vspace{-1.0em}
\caption{
\textbf{Data statistics of MA-Bench.} Action category definitions are consistent with the Micro-Action-52 dataset~\cite{guo2024benchmarking}.
}
\vspace{-1.5em}
\label{fig:dataset_static}
\end{figure*} 

\section{MA-Bench}\label{sec:bench}
\subsection{Overview} 
MA-Bench consists of 1,000 carefully curated micro-action videos covering all 52 micro-action categories~\cite{guo2024benchmarking}.
These videos have an average duration of 2.12 seconds, with a maximum length of 5.01 seconds. 
Through the semi-automatic annotation pipeline, we construct 12,000 challenging question–answer (QA) pairs organized in a three-tier hierarchical architecture encompassing eight fine-grained micro-action understanding tasks. 
As shown in Figure~\ref{fig:dataset_static}, we present the dataset statistics of MA-Bench.
\textit{First}, both action-level and body-level categories exhibit a long-tailed distribution, which increases the challenge of understanding rarely occurring micro-actions. 
\textit{Second}, micro-actions are short in duration, with the majority lasting between 1 and 3 seconds, while only a few exceed 4 seconds.
\textit{Third}, the word cloud reveals that the benchmark primarily focuses on fine-grained hand and directional movements (\eg, ``right hand,'' ``leftward,'' ``upward''), accompanied by clear temporal segmentation (``early,'' ``phase,'' ``segment'') and reasoning-oriented language (``indicating,'' ``described,'' ``based'').
This indicates that MA-Bench emphasizes not only perceptual recognition of body-part dynamics but also structured temporal reasoning and interpretive understanding of micro-actions. 

\subsection{Data Curation}
We selected videos from the large-scale micro-action recognition dataset Micro-Action-52 (MA-52)~\cite{guo2024benchmarking}, which comprises 22, 422 video instances collected from 205 participants of diverse age groups during psychological interview sessions. 
Each video in MA-52 is annotated with both body-level and action-level categories. 
To enhance the diversity of action patterns in MA-Bench, we randomly selected videos from 20 participants to construct the MA-Bench subset. 
In addition, we also established MA-Bench-Train, which includes 20, 510 (20.5K) videos from 166 participants with fine-grained motion captions to enable the fine-tuning of multimodal large language models for micro-action understanding. 
It is worth noting that a cross-subject setting is maintained between MA-Bench-Train and MA-Bench to ensure non-overlapping participants.

\begin{figure*}[t!]
\centering
\includegraphics[width=1\linewidth]{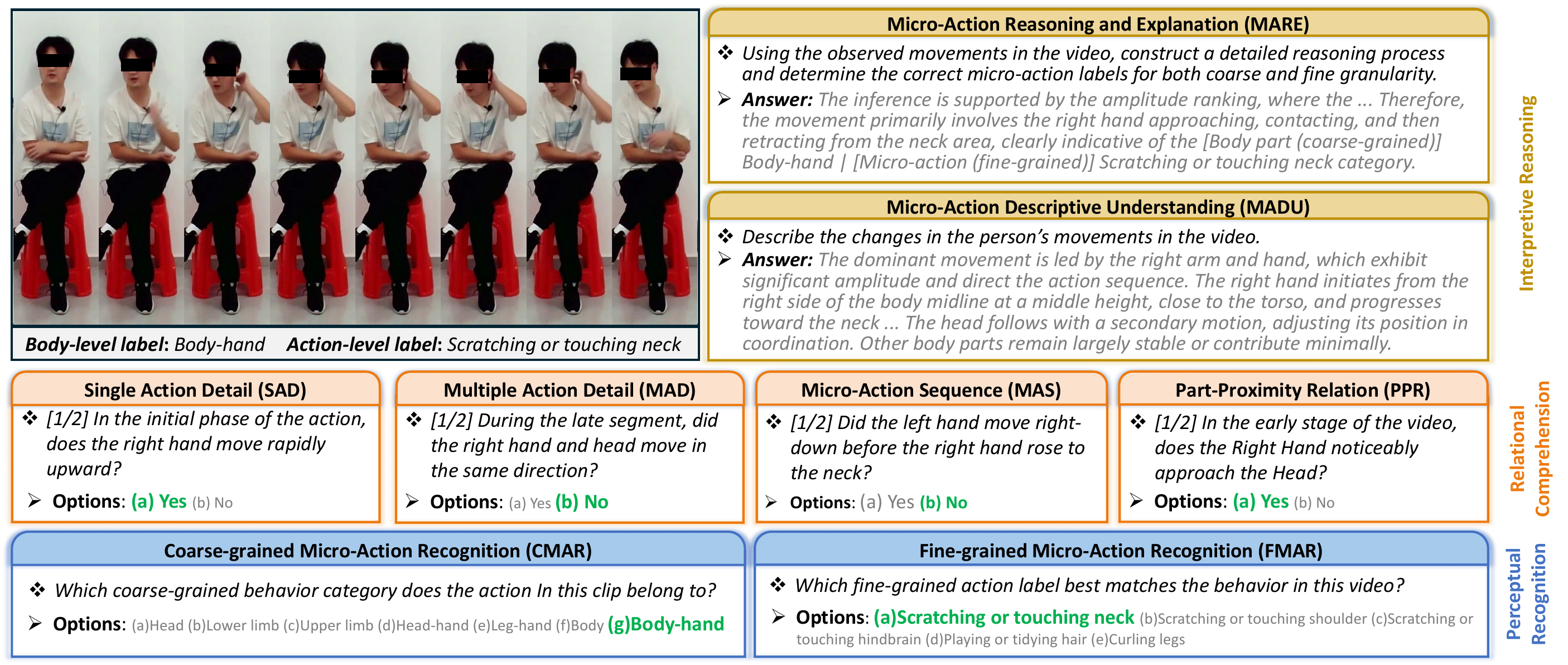}
\caption{
\textbf{Example of question-answer pairs from the proposed MA-Bench.}}
\label{fig:dataset}
\end{figure*}

\subsection{Task Definition}
Inspired by previous work in fine-grained motion understanding~\cite{hong2025motionbench,tu2025favor}, we design a three-tier evaluation architecture, termed ``\textbf{Perception–Comprehension–Reasoning}'', to assess the capability of MLLMs in fine-grained micro-action understanding, ranging from basic recognition to interpretive reasoning.

\ding{182} \textbf{Perceptual Recognition.} This requires the model to perform hierarchical categorization of micro-actions, identifying both coarse-grained and fine-grained action classes to establish a foundational perceptual understanding~\cite{guo2024benchmarking,li2025prototypical,gu2025motion}. The two tasks are defined as follows:

\noindent\ding{118} \textbf{\textit{Coarse-grained Micro-Action Recognition (CMAR)}.}
This task parallels video action recognition but focuses on body-level comprehension of subtle human movements. It aims to capture global motion patterns and overall body-region dynamics that characterize micro-actions, without delving into local or part-specific details.

\noindent\ding{118} \textbf{\textit{Fine-grained Micro-Action Recognition (FMAR)}.}
In contrast, this task targets action-level discrimination of micro-actions. It emphasizes fine spatial–temporal variations and nuanced motion cues that differentiate closely related micro-actions, reflecting the model’s ability to perceive subtle distinctions in human behavior. 

\ding{183} \textbf{Relational Comprehension.}
Compared to perceptual recognition, this task requires the model to focus on spatial-temporal reasoning and inter-part dynamics, examining how different body parts interact and coordinate during subtle movements. The question-answer pairs are organized in YES/NO format, with each pair targeting a distinct aspect of structural or temporal reasoning. 

\noindent\ding{118} \textbf{\textit{Single Action Detail (SAD).}}
This task assesses whether the model can accurately interpret how a specific atom action is executed, focusing on details (\eg, motion direction, amplitude, and rhythm). It evaluates fine-grained sensitivity to local motion characteristics. 

\noindent\ding{118} \textbf{\textit{Micro-Action Sequence (MAS).}}
This task evaluates the model’s ability to capture the temporal evolution of subtle micro-actions. It requires identifying the correct order or transition of local movements, reflecting an understanding of dynamic dependencies across time.

\noindent\ding{118} \textbf{\textit{Multiple Micro-Action Detail (MMAD).}}
This task evaluates the model’s ability to infer relational attributes among multiple body-part movements within a video, focusing on their temporal dependencies, such as simultaneity, transition, and directional consistency. It aims to assess higher-level reasoning that captures the integrated dynamics of whole-body behavior rather than isolated local motions.

\noindent\ding{118} \textbf{\textit{Part-Proximity Relation (PPR)}.}
This task evaluates the model’s ability to perceive and reason about spatial proximity changes between two body parts over time.
By focusing on relational spatial dynamics rather than individual motion patterns, it tests fine-grained understanding of body-part interaction and coordination within the micro-action context.

\ding{184} \textbf{Interpretive Reasoning.}
Compared to \ding{182} and \ding{183}, this requires model to generate detailed motion descriptions and reasoning-based classifications, bridging perception and explanation through natural-language interpretation.

\noindent\ding{118} \textbf{\textit{Micro-Action Descriptive Understanding (MADU).}}
This task requires the model to objectively describe the observed body movements in the video without performing reasoning or inference. The model should clearly depict the body parts involved, the type and manner of motion, and other observable physical details, forming a concise and accurate description of the ongoing micro-actions.

\noindent\ding{118} \textbf{\textit{Micro-Action Reasoning and Explanation (MARE).}}
This task requires generating a coherent reasoning chain that links analytical evidence to labeling decisions. The model should describe the micro-action, the corresponding coarse- and fine-grained micro-action labels, and briefly explain the reasoning chains for each label.

\begin{figure}[t!]
\centering
\includegraphics[width=1.0\linewidth]{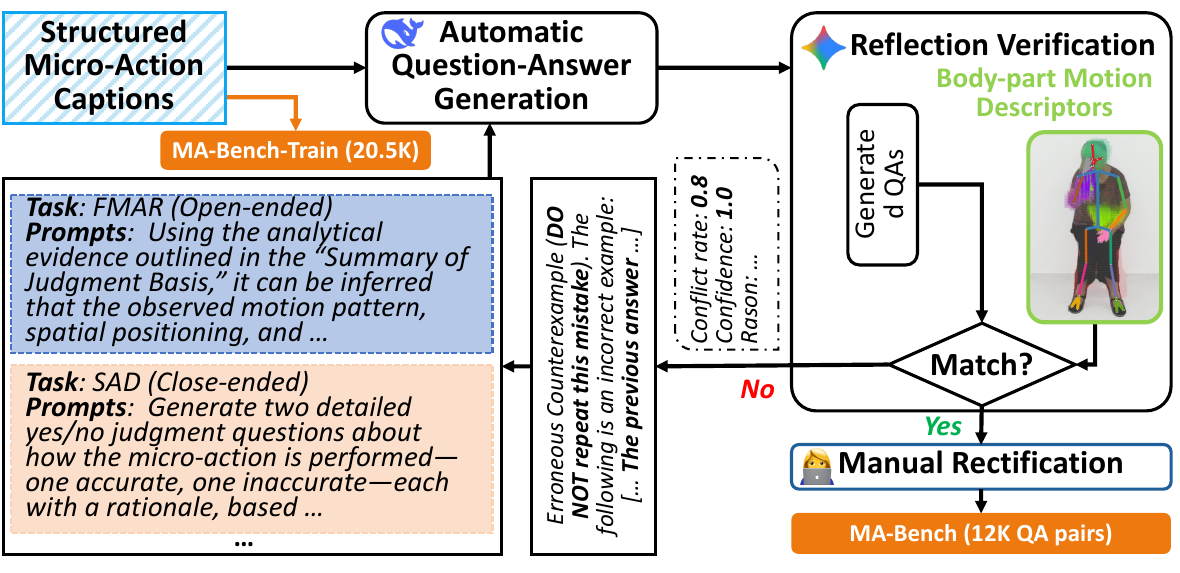}
\caption{\textbf{The pipeline for semi-automatic question-answer generation.}}
\vspace{-1.5em}
\label{fig:qa_gen}
\end{figure}

\subsection{Question-Answer Generation}
The source videos are mainly selected from MA-52~\cite{guo2024benchmarking} and are annotated only with action-level and body-level labels. \textit{However}, these annotations are insufficient for achieving fine-grained understanding of human micro-actions, as they lack explicit descriptions of motion patterns. 
To bridge this gap, we design \ding{182} \textbf{Structured Micro-Action Captions Generation} to generate structured micro-action motion captions and \ding{183} \textbf{Semi-automatic QA Generation} to construct MA-Bench and MA-Bench-Train.

\begin{table*}[t!]
\centering
\tabcolsep 4pt
\renewcommand{\arraystretch}{1.1}
\caption{The overall performances of 21 MLLMs on MA-Bench, including proprietary and open-source models. ``Acc.'' denotes the accuracy metric. ``AVG'' means the average score of all the tasks.  
The best results among all MLLMs are indicated in \textbf{bold}. 
}
\resizebox{1.0\linewidth}{!}{
\begin{tabular}{|l|c|cc|cccc|c|ccc|ccc|c|}
\hline\thickhline
\rowcolor{mygray}  &  &  \multicolumn{7}{c|}{\textbf{Closed-ended}} & \multicolumn{7}{c|}{\textbf{Open-ended}} \\ \cline{2-16}
\rowcolor{mygray} & & CMAR & FMAR & SAD & MAD & MAS & PPR  &   & \multicolumn{3}{c|}{MADU} & \multicolumn{3}{c|}{MARE} &   \\
\rowcolor{mygray} \multirow{-3}{*}{\textbf{Methods}} & \multirow{-3}{*}{\textbf{Date}}  & Acc. & Acc. & Acc. & Acc. & Acc. & Acc. & \multirow{-2}{*}{AVG} & L1 & L2 & L3 & L1 & L2 & L3 & \multirow{-2}{*}{AVG} \\
\hline
Full mark & -- & 100 & 100 & 100 & 100 & 100 & 100  & 100 & 5 & 5 & 5 & 5 & 5 & 5 & 5 \\
Random & -- & 14.7  & 20  & 50  & 50  & 50  & 50    & 39.05 & -- & -- & -- & -- & -- & -- & -- \\

\hline
\multicolumn{16}{|l|}{\textcolor{gray}{{\textit{\textbf{Proprietary MLLMs}}}}}\\ 
GPT-4o~\cite{hurst2024gpt} & 2024-11 & 20.50 & 30.70 & 51.30 & \textbf{62.35} & 49.25 & 55.10 & 44.87 & 0.60 & \textbf{0.65} & \textbf{0.54} & 1.02 & \textbf{0.91} & \textbf{0.67} & 0.73 \\

Gemini-2.5-Flash~\cite{comanici2025gemini} & 2025-06   & \textbf{43.00} & \textbf{31.40} & \textbf{56.55} & 60.50 & \textbf{55.50} & \textbf{57.25} & \textbf{50.70} & \textbf{0.63} & 0.62 & \textbf{0.54} & \textbf{1.22} & 0.86 & 0.71 & \textbf{0.76}   \\

Grok-2-vision~\cite{grok2} & 2024-12 & 19.30 & 31.30 & 52.85 & 54.25 & 47.75 & 54.05 & 43.25 & 0.51 & 0.50 & 0.41 & 1.02 & 0.80 & 0.63 & 0.65  \\
\hline
    
\multicolumn{16}{|l|}{\textcolor{gray}{{\textit{\textbf{Open-source MLLMs}}}}}\\

LLaVA-NeXT-Video-7B~\cite{zhang2024llavanextvideo}  & 2024-06 & 12.00 & 18.80 & 49.50 & 48.80 & 50.65 & 50.55 & 38.38 & 0.22 & 0.21 & 0.18 & 0.22 & 0.17 & 0.12 & 0.19 \\
VideoLLaMA2-7B~\cite{cheng2024videollama2}       & 2024-06 & 18.90 & 24.70 & 50.00 & 49.45 & 50.20 & 50.00 & 40.54 & 0.10 & 0.09 & 0.07 & 1.11 & 0.95 & 0.55 & 0.87 \\
InternVL2-8B~\cite{chen2024internvl2.5}      & 2024-06 & 17.40 & 23.90 & 51.40 & 49.60 & 51.25 & 53.15 & 41.12 & \textbf{0.58} & \textbf{0.58} & \textbf{0.50} & 0.68 & 0.43 & 0.34 & 0.52 \\
InternVideo2-Chat-8B~\cite{wang2024internvideo2} & 2024-08 & 22.90 & 28.10 & 57.60 & 58.95 & 55.80 & 49.00 & 45.39 & 0.12 & 0.08 & 0.06 & 0.04 & 0.03 & 0.02 & 0.06 \\
Phi-3.5-Vision~\cite{abdin2024phi3}      & 2024-08 & 16.20 & 23.70 & 51.25 & 53.00 & 50.20 & 44.40 & 39.79 & 0.26 & 0.26 & 0.23 & 0.21 & 0.16 & 0.11 & 0.21 \\
Pixtral-12B~\cite{agrawal2024pixtral}          & 2024-09 & 14.40 & 19.40 & 49.55 & 57.60 & 48.50 & 51.00 & 40.08 & 0.10 & 0.03 & 0.04 & 0.67 & 0.19 & 0.07 & 0.18 \\
LLaVA-OneVision-7B~\cite{li2024llava}   & 2024-09 & 18.50 & 26.00 & 50.30 & 53.60 & 54.35 & 50.85 & 42.27 & 0.39 & 0.38 & 0.33 & 0.25 & 0.23 & 0.16 & 0.29 \\
Qwen2-VL-7B~\cite{yang2024qwen2technicalreport} & 2024-09 & 28.50 & 32.00 & 53.15 & 57.10 & 52.75 & 60.25 & 47.29 & 0.39 & 0.32 & 0.28 & 0.35 & 0.29 & 0.21 & 0.31 \\
H2OVL Mississippi-2B~\cite{galib2024h2ovl} & 2024-10 & 17.70 & 19.20 & 48.55 & 48.20 & 52.75 & 50.10 & 39.42 & 0.14 & 0.11 & 0.08 & 0.18 & 0.10 & 0.06 & 0.11 \\
VideoLLaMA2.1-7B~\cite{cheng2024videollama2} & 2024-10 & 34.30 & 26.80 & 52.50 & 42.40 & 50.95 & 48.45 & 42.57 & 0.11 & 0.09 & 0.07 & 0.31 & 0.25 & 0.17 & 0.17 \\
InternVL2.5-8B~\cite{chen2024internvl2.5}       & 2024-12 & 19.10 & 27.10 & 51.70 & 47.00 & 47.90 & 49.80 & 40.43 & 0.48 & 0.49 & 0.41 & 0.65 & 0.45 & 0.35 & 0.47 \\
VideoChat-Flash-Qwen2-7B~\cite{li2024videochatflash} & 2025-01 & \textbf{36.90} & 33.60 & 54.25 & 64.10 & 49.80 & \textbf{60.55} & \textbf{49.87} & 0.09 & 0.07 & 0.05 & 0.08 & 0.06 & 0.03 & 0.06 \\
InternVideo2.5-Chat-8B~\cite{wang2025internvideo2}   & 2025-01 & 22.90 & 28.10 & 57.60 & 58.95 & 55.80 & 49.00 & 45.39 & 0.11 & 0.07 & 0.06 & 0.04 & 0.03 & 0.01 & 0.05 \\
VideoLLaMA3-2B~\cite{zhang2025videollama3}       & 2025-01 & 18.60 & 29.60 & 51.30 & 60.90 & 47.60 & 51.20 & 43.20 & 0.31 & 0.15 & 0.12 & 0.05 & 0.05 & 0.02 & 0.12 \\
VideoLLaMA3-7B~\cite{zhang2025videollama3}       & 2025-01 & 16.40 & 30.80 & \textbf{58.60} & \textbf{67.60} & \textbf{57.30} & 60.20 & 48.48 & 0.24 & 0.10 & 0.09 & 0.09 & 0.07 & 0.03 & 0.10 \\
InternVL3-8B~\cite{zhu2025internvl3} & 2025-04 & 17.10 & 31.20 & 50.10 & 57.40 & 51.85 & 54.80 & 43.74 & 0.15 & 0.09 & 0.05 & 0.15 & 0.11 & 0.06 & 0.10 \\
Qwen2.5-VL-3B~\cite{yang2024qwen2.5}        & 2025-01 & 16.90 & 25.80 & 51.05 & 53.20 & 47.60 & 56.40 & 41.83 & 0.28 & 0.25 & 0.22 & 0.41 & 0.32 & 0.23 & 0.29 \\
Qwen2.5-VL-7B~\cite{yang2024qwen2.5}        & 2025-01 & 20.40 & 29.20 & 53.50 & 56.45 & 48.30 & 54.40 & 43.71 & 0.37 & 0.37 & 0.30 & 0.49 & 0.41 & 0.30 & 0.37 \\
Phi-4-Multimodal~\cite{abdin2024phi4}     & 2025-03 & 11.40 & 25.60 & 56.40 & 50.85 & 54.55 & 45.05 & 40.64 & 0.15 & 0.09 & 0.05 & 0.15 & 0.11 & 0.06 & 0.10 \\
Qwen3-VL-8B~\cite{qwen3technicalreport} & 2025-10 & 32.23 & \textbf{37.11} & 56.10 & 53.69 & 47.94 & 54.74 & 46.97 & 0.50 & 0.51 & 0.47 & \textbf{1.18} & \textbf{1.00} & \textbf{0.84} & \textbf{0.75} \\ \hline
\rowcolor{mygray} Qwen3-VL-8B~\cite{qwen3technicalreport} + MA-Bench-Train & 2025-10 & \textbf{47.90} &  32.60 & \textbf{60.30} &  59.65 & 50.00 & 53.60 & \textbf{50.68} & \textbf{1.50} &  \textbf{1.67} & \textbf{1.54} & \textbf{1.98} & \textbf{1.78} & \textbf{1.67} & \textbf{1.69} \\ \hline
\end{tabular}}
\vspace{-0.5cm}
\label{tab:ourbench}
\end{table*}

\ding{182} \textbf{Structured Micro-Action Captions Generation.}
As illustrated in Figure~\ref{fig:teaser}, the micro-motion tracker is designed to construct body-part motion descriptors for each video. Specifically, it first employs CoTracker3~\cite{karaev2025cotracker3} to extract dense optical flow with four directional components that capture detailed motion dynamics, and then utilizes the YOLOv8x-Seg model~\cite{varghese2024yolov8} to segment human-centric regions.
Finally, we align the segmented optical flow with the skeleton data~\cite{gu2025motion} to generate body-part motion descriptors that summarize both motion vectors and spatial coordinates for each body part. 
Building upon these descriptors, we prompt the large language models (LLMs, \eg, DeepSeek-v3.2~\cite{deepseekai2024deepseekv32}, DeepSeek-R1~\cite{deepseekai2025deepseekr1incentivizingreasoningcapability}) to summarize the motion patterns of individual body parts, producing body-part motion captions. Subsequently, we prompt LMMs to synthesize these descriptions into a structured micro-action caption that involves the fine-grained motion details. 
These captions are used to build the \textbf{MA-Bench-Train}. 
Detailed implementations are given in the supplementary materials. 

\ding{183} \textbf{Semi-automatic QA Generation.} 
Based on the generated structured micro-action captions, we design a three-stage semi-automatic pipeline to construct \textbf{MA-Bench}.  
As shown in Figure~\ref{fig:qa_gen}, the pipeline consists of automatic question-answer generation, reflection verification, and manual rectification. 
After the above steps, MA-Bench constructs an average of 2.9 multiple-choice questions for each video. 
\textbf{1) Automatic Question-Answer Generation.} 
Here, we automatically generate question–answer pairs using fine-grained motion captions as textual input. 
The LLM is prompted with structured templates that correspond to the pre-defined tasks. 
The LLM reformulates motion descriptions into multiple-choice or binary QA formats, ensuring that each question targets a distinct perceptual or reasoning dimension. 
\textbf{2) Reflection Verification.} 
To further ensure quality and correctness, the generated QA pairs are reviewed by an LLM serving as an automated validator. It cross-checks the alignment between the QA content and the motion descriptors and verifies whether each question–answer pair is logically consistent with the underlying visual evidence.
\textbf{3) Manual Rectification.}
Finally, human annotators conduct multi-round verification and correction to ensure factual and semantic accuracy. Annotators review all QA pairs, confirm that the questions are unambiguous, the options are mutually exclusive, and the correct answer is clearly grounded in the video evidence.

\section{Experiments}
\label{sec:exps}

\begin{table*}[t!]
\tabcolsep 8pt
\caption{\textbf{Ablation results of frames on Qwen3-VL-8B model.}}
\label{tab:abl_frms}
\resizebox{1.0\linewidth}{!}{
\begin{tabular}{|c|c|cc|cccc|c|ccc|ccc|c|}
\hline\thickhline
\rowcolor{mygray} & &  \multicolumn{7}{c|}{\textbf{Closed-ended}} & \multicolumn{7}{c|}{\textbf{Open-ended}} \\ \cline{3-16}
\rowcolor{mygray} & & CMAR & FMAR & SAD & MAD & MAS & PPR  & & \multicolumn{3}{c|}{MADU} & \multicolumn{3}{c|}{MARE} & \\
\rowcolor{mygray}
\multirow{-3}{*}{\textbf{Exp.}}
& 
\multirow{-3}{*}{\textbf{\#Frames}}
& Acc & Acc & Acc & Acc & Acc & Acc & \multirow{-2}{*}{AVG} & L1 & L2 & L3 & L1 & L2 & L3 & \multirow{-2}{*}{AVG} \\ \hline
A1 & 4  & 31.73 & 36.64 & \textbf{56.11} & 53.50 & \textbf{47.85} & 53.83 & 46.61 & 0.49 & 0.51 & 0.46 & 1.09 & 0.93 & 0.77 & 0.71\\
A2 & 8  & \textbf{32.23} &
\textbf{37.11} &
56.10 &
\textbf{53.69} &
\textbf{47.94} &
54.74 &
\textbf{46.97} & \textbf{0.50} & \textbf{0.51} & \textbf{0.47} & \textbf{1.18} & \textbf{1.00} & \textbf{0.84} & \textbf{0.75}\\ 
A3 & 16 & 31.49 & 36.11 & 56.02 & 53.71 & 47.94 & \textbf{54.96} & 46.71 & 0.48 & 0.49 & 0.44 & 1.13 & 0.98 & 0.82 & 0.72 \\ \hline
\end{tabular}}
\end{table*}

\begin{table*}[t!]
\caption{\textbf{Supervised Fine-tuning results on MA-Bench using MA-Bench-Train.}}
\tabcolsep 4pt
\renewcommand{\arraystretch}{1.1}
\resizebox{1.0\linewidth}{!}{
\begin{tabular}{|c|l|c|cc|cccc|c|ccc|ccc|c|}
\hline\thickhline
\rowcolor{mygray} & &  &  \multicolumn{7}{c|}{\textbf{Closed-ended}} & \multicolumn{7}{c|}{\textbf{Open-ended}} \\ \cline{3-17}
\rowcolor{mygray} & & & CMAR & FMAR & SAD & MAD & MAS & PPR  & & \multicolumn{3}{c|}{MADU} & \multicolumn{3}{c|}{MARE} &   \\
\rowcolor{mygray} \multirow{-3}{*}{\textbf{Exp.}} & \multirow{-3}{*}{\textbf{Methods}} & \multirow{-3}{*}{\textbf{SFT (LoRA)}}  & Acc. & Acc. & Acc. & Acc. & Acc. & Acc. & \multirow{-2}{*}{AVG} & L1 & L2 & L3 & L1 & L2 & L3 & \multirow{-2}{*}{AVG} \\ \hline
B1 & Qwen3-VL-8B~\cite{qwen3technicalreport} & - & 32.23 &  \textbf{37.11} &  56.10 &  53.69 & 47.94 & \textbf{54.74} & 46.97 & 0.50 &  0.51 & 0.47 & 1.18 & 1.00 & 0.84 & 0.75\\ \hline
B2 & + MA-Bench-Train & Vision Encoder & 41.20 & 33.50 & 58.20 & 56.90 &  49.10 & 52.80 & 48.62 & 0.95 & 1.05 & 0.98 & 1.35 & 1.20 & 1.10 & 1.11 \\ 
B3 & + MA-Bench-Train & LM Decoder & \textbf{49.10} & 34.70 & 58.65 &  58.80 & 50.30 &  52.25 & 50.63 & 1.41 & 1.57 & 1.45 & 1.91 & 1.69 & 1.56 & 1.60 \\ 
B4 & + MA-Bench-Train & All & 47.90 &  32.60 & \textbf{60.30} & \textbf{59.65} & 50.00 & 53.60 & \textbf{50.68} & \textbf{1.50} &  \textbf{1.67} & \textbf{1.54} & \textbf{1.98} & \textbf{1.78} & \textbf{1.67} & \textbf{1.69} \\ \hline
\end{tabular}}
\vspace{-1.5em}
\label{tab:fine-tuning}
\end{table*}

\subsection{Experimental Setup}
We evaluate 23 released MLLMs, covering a wide range of model sizes, architectures, and training methodologies. For proprietary MLLMs, we evaluate GPT-4o~\cite{hurst2024gpt}, Gemini-2.5-Flash~\cite{gemini2_5}, and Grok-2-Vision~\cite{grok2}. 
For open-sourced MLLMs, we evaluate Llava-NeXT-Video~\cite{zhang2024llavanextvideo}, VideoLLama2~\cite{cheng2024videollama2}, VideoLLama3~\cite{zhang2025videollama3}, InternVL2.5~\cite{chen2024internvl2.5},  InternVideo2~\cite{wang2024internvideo2}, InternVideo2.5~\cite{wang2025internvideo2}, 
VideoChat~\cite{li2024videochatflash},
H2OVL-Mississippi~\cite{galib2024h2ovl}, 
Phi-3.5-vision~\cite{abdin2024phi3}, Phi-4-multimodal~\cite{abdin2024phi4}, Llava-One-Vision~\cite{li2024llava}, Qwen2-VL~\cite{wang2024qwen2}, Qwen2.5-VL~\cite{yang2024qwen2.5}, and Qwen3-VL~\cite{qwen3technicalreport}. 
Following the common practice~\cite{li2024mvbench}, the valuations were conducted in a zero-shot setting, where the input comprised a video or a set of sampled frames followed by a prompt. 
The model was instructed to provide the final answer immediately, without any intermediate reasoning. The sample frames are set to 8 for each video. 
For closed-ended tasks, performance is measured using accuracy. 
For open-ended tasks, we adopt the LLM-as-a-judge protocol~\cite{zheng2023judging,zhao2025mmvu}, using GPT-4o~\cite{hurst2024gpt} as the evaluator. For micro-action descriptive understanding (MADU), we evaluate outputs along three dimensions: L1 measures the correctness of core action semantics, L2 estimates the accuracy of spatial directions and relations, and L3 considers the consistency of temporal order and structure. For micro-action reasoning and explanation (MARE), we use three dimensions: L1 measures the correctness of the coarse-grained body-level label, L2 assesses the accuracy of the fine-grained action-level label, and L3 examines the consistency of the causal reasoning chain supporting the prediction. Both MADU and MARE are scored on a 0–5 scale. 
The prompts used for the LLM judge are provided in the supplementary material.

\subsection{Performance Analysis}
The comprehensive evaluation on the proposed MA-Bench is reported in Table~\ref{tab:ourbench}, which systematically examines the fine-grained micro-action understanding capabilities of modern proprietary and open-source MLLMs.

\textbf{Closed-ended Evaluation.}
Proprietary models hold a clear advantage due to their larger training corpus and stronger multimodal alignment. Specifically, Gemini-2.5-Flash~\cite{comanici2025gemini} achieves the best average accuracy of 50.70\%, followed by GPT-4o~\cite{hurst2024gpt}, which obtains 44.87\%. 
Among open-source models, VideoChat-Flash-Qwen2-7B~\cite{li2024videochatflash} and InternVideo2.5-8B~\cite{wang2025internvideo2} demonstrate competitive performance, reaching 49.87\% and 44.09\% respectively, outperforming most other open-source baselines such as LLaVA-NeXT-Video-7B~\cite{zhang2024llavanextvideo} and Phi-3.5-Vision~\cite{abdin2024phi3}. 
After fine-tuning on the proposed MA-Bench-Train, Qwen3-VL-8B~\cite{qwen3technicalreport} further improves to 50.68\%, surpassing Gemini-2.5-Flash and becoming the best model overall, which confirms the strong transferability of MA-Bench-Train in enhancing fine-grained micro-action understanding. 

\textbf{Open-ended Evaluation.}
Compared to the closed-ended evaluation, the performance gap among models is larger. Proprietary models again lead, but their advantage is less dominant in reasoning-heavy tasks. 
For example, such as Gemini-2.5-Flash only achieves 0.76 on average.
In contrast, Qwen3-VL-8B, fine-tuned on the MA-Bench-Train, achieves the highest average score of 1.69, substantially outperforming all open-source counterparts and even proprietary models. These results suggest that while large proprietary MLLMs possess general multimodal capabilities, fine-grained micro-action reasoning benefits significantly from domain-adaptive supervision. 
The improvements across both closed- and open-ended tracks indicate that MA-Bench-Train not only enhances visual discrimination but also strengthens the model’s interpretative and descriptive ability for subtle human motions. 
In summary, open-source MLLMs fine-tuned with domain-specific supervision can achieve or even surpass the performance of proprietary models on micro-action understanding, highlighting the value of MA-Bench-Train for advancing research in fine-grained multimodal reasoning.

\begin{figure*}[t!]
\centering
\includegraphics[width=1.0\linewidth]{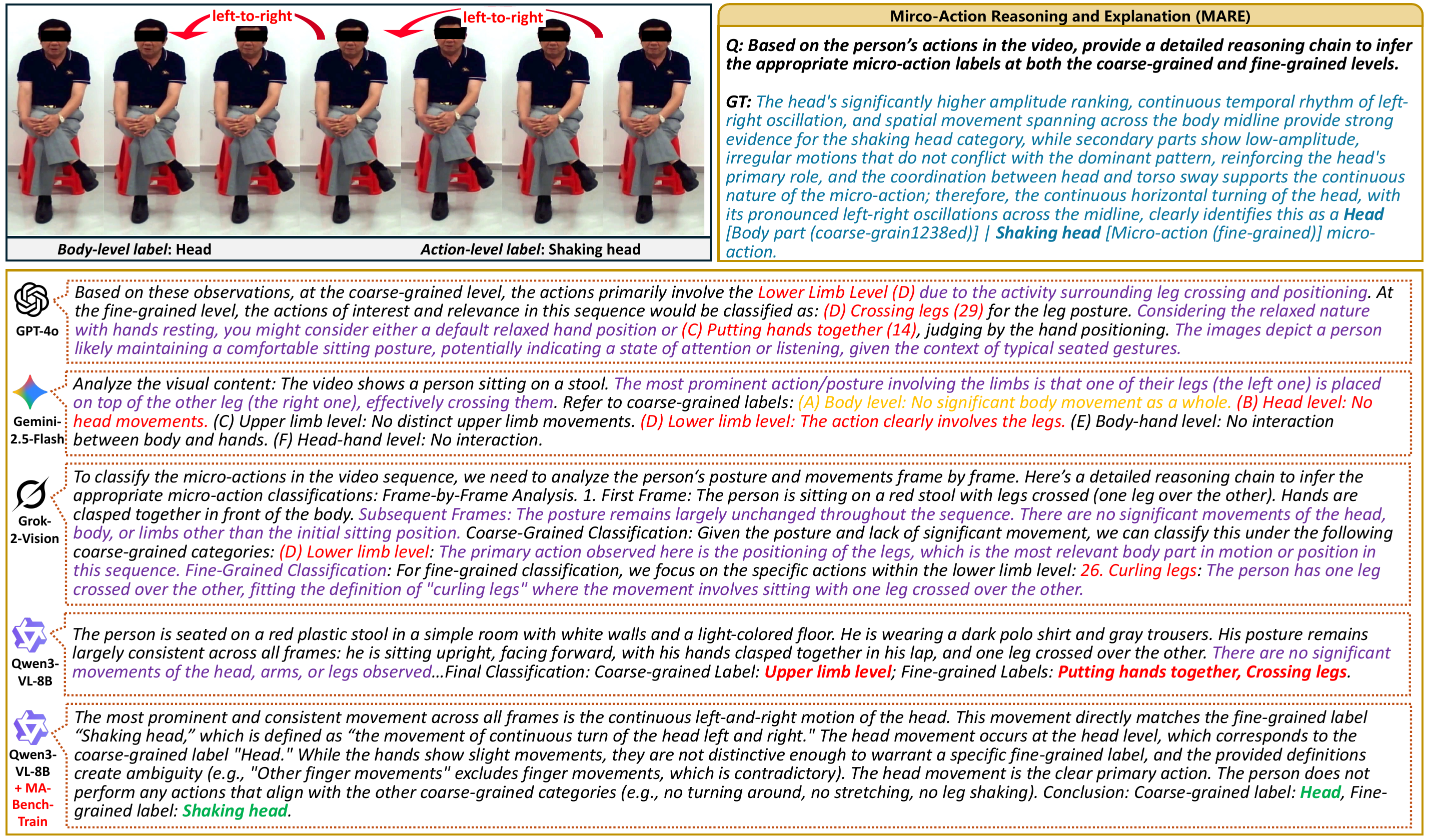}
\caption{
\textbf{Qualitative example on the task of Micro-Action Reasoning and Explanation (MARE).} Green font denotes correct predictions, red indicates completely incorrect ones, yellow marks partial errors, black shows results with no impact, and purple highlights faulty reasoning chains. 
More examples of other tasks are provided in the supplementary material.
}
\vspace{-1.5em}
\label{fig:vis_qual}
\end{figure*}

\section{Further Analysis}\label{sec:further_ana}

\textbf{The Influence of Video Frames.}
As shown in Table~\ref{tab:abl_frms}, the Qwen3-VL-8B model achieves the best overall performance when sampling 8 frames, with an average accuracy of 46.97\% on closed-ended tasks and 0.75 on open-ended tasks. Increasing the number of frames from 4 to 8 consistently improves most tasks, suggesting that moderate temporal redundancy enhances motion perception and reasoning. 
\textit{However}, further increasing the frame to 16 yields no additional gains and even causes a slight drop, indicating that excessive frame inputs may introduce visual redundancy and distract temporal attention. 
In summary, the 8-frame configuration provides an optimal trade-off between temporal coverage and computational efficiency.

\noindent \textbf{The Influence of Supervised Fine-tuning.} 
As introduced in Section~\ref{sec:intro}, we constructed MA-Bench-Train to enhance MLLMs' capability in fine-grained micro-action understanding. 
In this section, we fine-tune Qwen3-VL-8B via LLaMA-Factory~\cite{zheng2024llamafactory} and LoRA~\cite{hu2022lora}, and compare the performance before and after supervised fine-tuning. 
The experimental results are reported in Table~\ref{tab:fine-tuning}. 
Compared with the original model (Exp. B1), the fine-tuned models show consistent improvements on both closed-ended and open-ended tasks. 
Specifically, only fine-tuning the vision encoder (Exp. B2) or LM decoder (Exp. B3) yields a noticeable average gain on closed-ended tasks and open-ended ones. 
When fine-tuning both the whole module (Exp. B4), further improvements are achieved, reaching the best average scores of 50.68 (closed-ended) and 1.69 (open-ended). These results demonstrate that MA-Bench-Train effectively enhances the model’s fine-grained perception and multimodal reasoning capability. More implementation details 
% about these experiments 
can be found in the supplementary material.

\subsection{Qualitative Analysis}\label{sec:qual_vis}

To further examine the capabilities and limitations of frontier models on MA-Bench, we conduct comprehensive case studies on the task of Micro-Action Reasoning and Explanation (MARE). 
Limited to the page, additional qualitative analyses of other tasks are provided in the supplementary material. 
As illustrated in Figure~\ref{fig:vis_qual}, models not fine-tuned on MA-Bench-Train (\ie, zero-shot models) tend to misclassify or overlook subtle motion cues, frequently confusing head- and limb-related micro-actions or generating inconsistent reasoning chains. Although Qwen3-VL-8B shows partial improvement, its explanations remain somewhat ambiguous. 
In contrast, it fine-tuned on MA-Bench-Train produces more accurate and coherent reasoning, correctly identifying both the coarse-grained (Head) and fine-grained (Shaking head) actions with a clear and interpretable reasoning chain. 
These results reveal that current MLLMs are still struggling to capture subtle micro-action cues and fine-grained temporal dynamics.

\section{Conclusion}
We present MA-Bench, the first comprehensive benchmark for evaluating Multimodal Large Language Models (MLLMs) in the context of fine-grained micro-action understanding, consisting of 1,000 videos with both open-ended (10,000 multiple-choice QA pairs across 6 tasks) and closed-ended evaluation (2,000 QAs). 
The comprehensive evaluation of 23 modern MLLMs reveals substantial limitations in capturing the subtle, fine-grained temporal dynamics characteristic of micro-actions. 
To address this gap, we further construct MA-Bench-Train, comprising 20,510 annotated videos with detailed and structured micro-action captions, which leads to significant improvements in micro-action understanding performance on MA-Bench. 
We hope that the proposed MA-Bench and MA-Bench-Train will pave the way for future research to explore new methodologies for modeling subtle micro-action cues, thereby enhancing fine-grained micro-action understanding and enabling application in real-world micro-behavior analysis scenarios.

\section*{Acknowledgments}
This work is supported by National Key R\&D Program of China (2024YFB3311600), National Natural Science Foundation of China (62272144, 62502447, 92570101), the Anhui Provincial Natural Science Foundation (2408085J040), the Major Project of Anhui Provincial Science and Technology Breakthrough Program (202423k09020001), the UAEU Start-Up Research Grant G00005318, and the UAEU-ZU Joint Research Program G00005305.

{
\small
\bibliographystyle{ieeenat_fullname}
\bibliography{main}
}

% WARNING: do not forget to delete the supplementary pages from your submission 
% \input{sec/X_suppl}

\end{document}